\documentclass{article}
\usepackage{spconf,amsmath,graphicx}

\usepackage{enumitem}
\setlist{nosep, leftmargin=14pt}

\usepackage{mwe} % to get dummy images

\usepackage{amsmath,amssymb,amsfonts}
\usepackage[table]{xcolor} % 加载 xcolor 包并指定 table 选项
\usepackage{booktabs} % 加载 booktabs 包
\usepackage{multirow} % 加载 multirow 包
\usepackage{rotating} % 加载 rotating 包，用于旋转单元格内容
\usepackage[caption=false]{subfig}

\usepackage{fancyhdr}
\fancypagestyle{firstpage}{
	\fancyhf{} % 清除所有页眉页脚
	\fancyfoot[L]{\footnotesize Accepted by IEEE International Symposium on 
	Biomedical Imaging (ISBI) 2026}
}

\fancypagestyle{firstpage}{
	\fancyhf{}
	
	\fancyfoot[L]{\footnotesize Accepted by IEEE International Symposium on 
	Biomedical Imaging (ISBI) 2026}
}

\title{Diffusion Model-Based Data Augmentation \\ for Enhanced Neuron 
	Segmentation}
\name{Liuyun Jiang$^{1,2}$, Yanchao Zhang$^{1,2}$, Jinyue Guo$^{1,3}$, Yizhuo 
Lu$^{1,2}$, Ruining Zhou$^{1,2}$, and Hua Han$^{1,2^*}$
}
\address{$^{1}$State Key Laboratory of Brain Cognition and Brain-inspired 
Intelligence Technology, Institute of \\ Automation, Chinese Academy of 
Sciences \\
	$^{2}$School of Future Technology, University of Chinese Academy of 
	Sciences \\
	$^{3}$School of Artificial Intelligence, University of Chinese Academy of 
	Sciences
}
\begin{document}
%\ninept
%
\maketitle
\thispagestyle{firstpage}
\begin{abstract}
Neuron segmentation in electron microscopy (EM) aims to reconstruct the 
complete neuronal connectome; however, current deep learning-based methods are 
limited by their reliance on large-scale training data and extensive, 
time-consuming manual annotations. Traditional methods augment the training set 
through geometric and photometric transformations; however, the generated 
samples 
remain highly correlated with the original images and lack structural 
diversity. To address this limitation, we propose a diffusion-based data 
augmentation framework capable of generating diverse and structurally plausible 
image–label pairs for neuron segmentation. Specifically, the framework employs 
a 
resolution-aware conditional diffusion model with multi-scale conditioning and 
EM resolution priors to enable voxel-level image synthesis from 3D masks. It 
further incorporates a biology-guided mask remodeling module that produces 
augmented masks with enhanced structural realism. Together, these components 
effectively enrich the training set and improve segmentation performance. On 
the AC3 and AC4 datasets under low-annotation regimes, our method improves the 
ARAND metric by 32.1\% and 30.7\%, respectively, when combined with two 
different post-processing methods. Our code is available at 
https://github.com/HeadLiuYun/NeuroDiff.
\end{abstract}
\section{Introduction}
Neuron segmentation aims to elucidate brain function by mapping neural 
connectivity and analyzing inter-neuronal signaling pathways. Advances in 
volume EM have enabled nanoscale reconstruction of 
three-dimensional neuronal structures \cite{em1,em2}; however, the increasing 
data scale and resolution pose significant challenges for automated processing. 
Deep learning-based segmentation methods rely heavily on large annotated 
datasets, making training both time-consuming and labor-intensive, while 
limited data diversity often results in overfitting and poor generalization.

\begin{figure}[t]
	\centering
	\includegraphics[width=0.95\columnwidth]{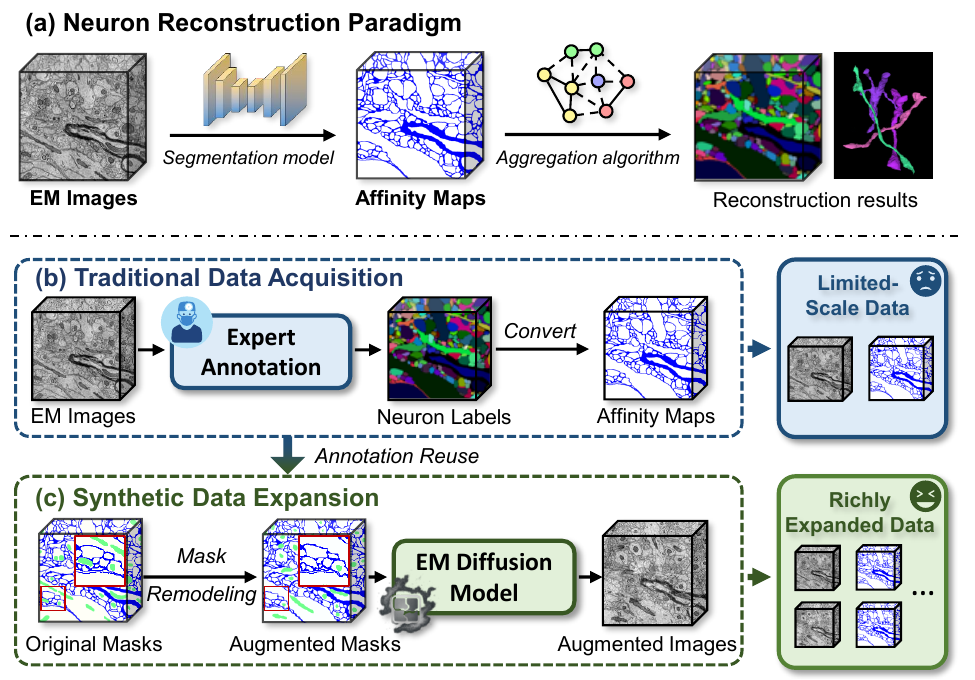} 
	\vspace{-0.3cm}
	\caption{(a) Illustration of the neuron reconstruction paradigm: EM images 
	are processed by a segmentation network to generate affinity maps, which 
	are aggregated by a post-processing algorithm to obtain the final 
	reconstruction. (b) Traditional data acquisition relies on manual 
	annotation for limited-scale training data. (c) Our method generates new 
	image–label pairs to enrich the training dataset.
	}
	\label{fig1}
	\vspace{-0.3cm}
\end{figure}

Data augmentation is a widely used technique to increase both the quantity and 
diversity of labeled data. Previous neuron segmentation 
methods~\cite{superhuman} mainly rely on basic geometric and photometric 
transformations (e.g., flipping, rotation, brightness adjustment), which 
provide limited diversity and produce augmented images similar to the 
originals. Recently, generative 
model-based augmentation strategies have shown promising results in biomedical 
imaging~\cite{MRDA,Medddpm,aug1}, yet they have not been extended to 3D neuron 
segmentation in EM images. A key challenge lies in the precise 
voxel-level control required for the elongated, slender morphology of neurons.

\begin{figure}[t]
	\centering
	\includegraphics[width=0.48\textwidth]{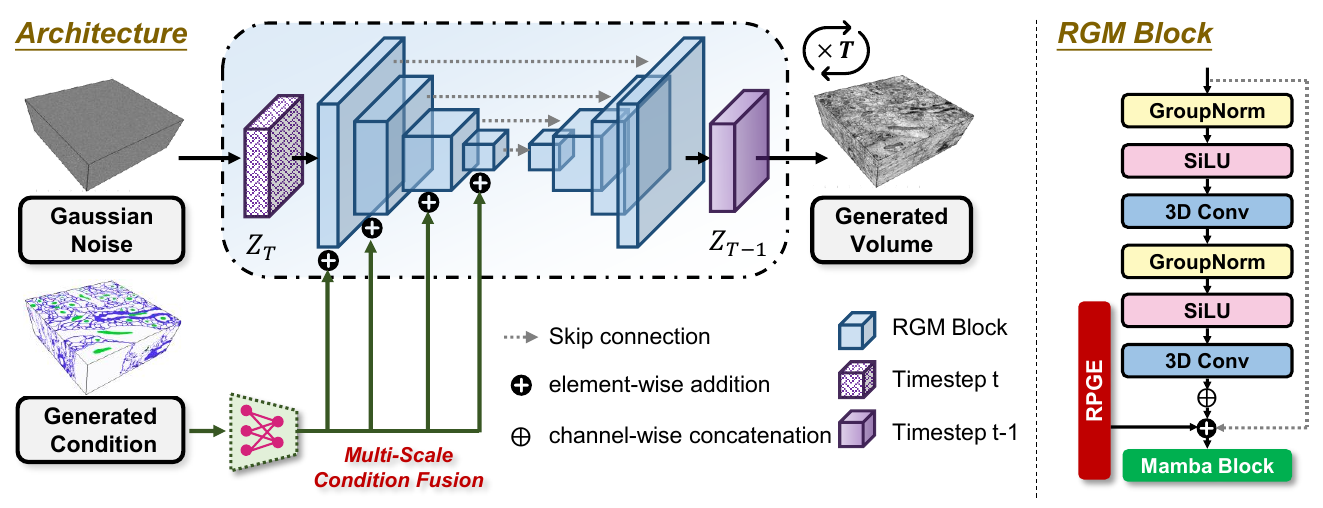} % Reduce the figure size so 
	%that it is slightly narrower than the column.
	\vspace{-0.80cm}
	\caption{Architecture of the proposed Resolution-Aware Conditional 
	Diffusion model.}
	\label{fig2}
	\vspace{-0.3cm}
\end{figure}
To address the scarcity of training data and the challenges of neuron instance 
segmentation, as illustrated in Fig.~\ref{fig1}, we propose a diffusion-based 
data augmentation framework for EM images. It improves segmentation performance 
by generating diverse and structurally plausible image–label pairs that enrich 
the training set. The framework comprises two key modules: a 3D EM image 
synthesis module and a 3D mask remodeling module. The synthesis module employs 
a conditional diffusion model to generate EM images from 3D masks, while the 
remodeling module produces augmented masks guided by boundary morphology, 
biological priors, and structural plausibility. By integrating both modules, 
our method synthesizes additional image–label pairs from existing data, 
effectively expanding the training set and enhancing segmentation performance.

The contributions of this work are summarized as follows: 
(1) We propose a diffusion-based data augmentation framework for neuron 
segmentation. 
(2) We integrate EM resolution priors into the diffusion model and incorporate 
biological priors and structural constraints into 3D mask remodeling to 
generate structurally consistent synthetic data. 
(3) Experimental results show that our method improves neuron segmentation 
performance and achieves state-of-the-art results, particularly under limited 
labeled data conditions.

\section{Method}

\subsection{Resolution-Aware Conditional Diffusion}
In this study, we employ a Denoising Diffusion Probabilistic Model 
(DDPM)~\cite{ddpm} for 3D EM image synthesis. 
Following previous 3D diffusion studies~\cite{3dddpm,Medddpm}, we extend the 
U-Net to a 3D architecture for volumetric generation. 
In the forward process, Gaussian noise $\epsilon \sim \mathcal{N}(0, I)$ is 
gradually added to the data $x_0$ over $T$ timesteps, where $\beta_t$ 
represents the variance of the added noise:
\begin{equation}
	q(x_{1:T} \mid x_0) = \prod_{t=1}^{T} \mathcal{N}\left(x_t; 
	\sqrt{1-\beta_t}\,x_{t-1},\, \beta_t I\right)
\end{equation}
In the reverse process, the model denoises the data from $x_T$ to reconstruct 
$x_0$. 
For conditional generation, the U-Net takes both the noisy input $x_T$ and the 
condition $c$, 
where $c$ integrates mitochondrial masks and neuronal boundaries, 
since mitochondrial membranes affect the identification of neuronal boundaries. 
The conditional reverse diffusion process is defined as:
\begin{equation}
	p_\theta(x_{t-1} \mid x_t, c)
	= \mathcal{N}\left(x_{t-1};\, \mu_\theta(x_t, t, c),\, \Sigma_\theta(x_t, 
	t, c)\right)
\end{equation}
The U-Net predicts the mean $\mu_\theta$ and variance 
$\Sigma_\theta$, while estimating the noise $\epsilon_\theta$ to reverse the 
diffusion process. New EM data $x_0$ are generated by iteratively sampling 
$x_{t-1} 
\sim p_\theta(x_{t-1} \mid x_t, c)$ for $t = T, \ldots, 1$.

To enable voxel-level control, our denoising network incorporates two essential 
modules: the Multi-Scale Conditioning (MSC) module and the Resolution-Prior 
Guided Global Modeling (RGM) module.

\noindent\textbf{Multi-Scale Conditioning Module.}
Different U-Net layers capture distinct semantic levels, and integrating 
conditions across multiple scales provides finer control. Therefore, we employ 
a multi-scale conditioning strategy rather than directly concatenating the 
condition $c$ with $x_T$. Moreover, neuronal and organelle structures in EM 
images exhibit considerable variations in both size and morphology. The 
proposed 
multi-scale conditioning mechanism enables the model to synthesize large-scale 
structures while preserving fine details.

Let the input condition be $c \in \mathbb{R}^{2 \times D \times H \times W}$, 
where $D$, $H$, and $W$ represent the spatial dimensions of depth, height, and 
width. We first project $c$ through several 3D convolutional layers to obtain a 
low-resolution, high-channel embedding $c_{\text{embed}} \in \mathbb{R}^{256 
	\times D \times \tfrac{H}{8} \times \tfrac{W}{8}}$. For each U-Net feature 
layer $f_i \in \mathbb{R}^{C_i \times D_i \times H_i \times W_i}$, the embedded 
feature $c_{\text{embed}}$ is resized and processed by a zero-initialized 
convolution to produce $c_i \in \mathbb{R}^{C_i \times D_i \times H_i \times 
	W_i}$. The aligned conditional feature is subsequently injected into the 
corresponding U-Net layer via element-wise addition: $f_i \leftarrow f_i + 
c_i$, as illustrated in Fig.~\ref{fig2}.

\noindent\textbf{Resolution-Prior Guided Global Modeling.}
Structures in EM images exhibit spatial continuity and connectivity, requiring 
the denoising network to capture global voxel dependencies for structurally 
consistent 3D generation. Mamba~\cite{mamba}, a backbone built upon 
state-space models (SSMs)~\cite{ssm}, offers linear computational complexity 
and excels in long-sequence modeling. Inspired by this property, we integrate 
Mamba into our denoising network to enable efficient and globally coherent 3D 
EM image synthesis.

Both Mamba and Transformers operate on serialized one-dimensional sequences. 
Vision Transformers~\cite{vit} partition images into fixed-size patches and 
flatten them into vectors, which overlook intra-patch voxel relationships and 
constrain voxel-level tasks such as neuron segmentation. Leveraging Mamba’s 
linear complexity, we directly flatten the 3D volume into a one-dimensional 
sequence to capture voxel-level dependencies without patch partitioning. To 
preserve 3D spatial context, we introduce a Resolution-Prior Guided Encoding 
(RPGE) module that incorporates the anisotropic resolution priors of EM images:
\begin{equation}
	\operatorname{RPGE}(z_i,y_i,x_i) = \operatorname{MLP}\!\left(\left[
	z_i \cdot r_z,\; y_i \cdot r_{xy},\; x_i \cdot r_{xy}
	\right]\right)
\end{equation}
where $r_z$ and $r_{xy}$ represent the axial and lateral resolutions of the EM 
volume, and $(z_i, y_i, x_i)$ denote the coordinates of the $i$-th voxel. As 
illustrated in Fig.~\ref{fig2}, the RPGE output is added element-wise within 
the RGM module and subsequently fed into the Mamba block for global modeling.

\subsection{Biology-Guided Mask Remodeling}
\begin{figure}[t]
	\centering
	\includegraphics[width=0.48\textwidth]{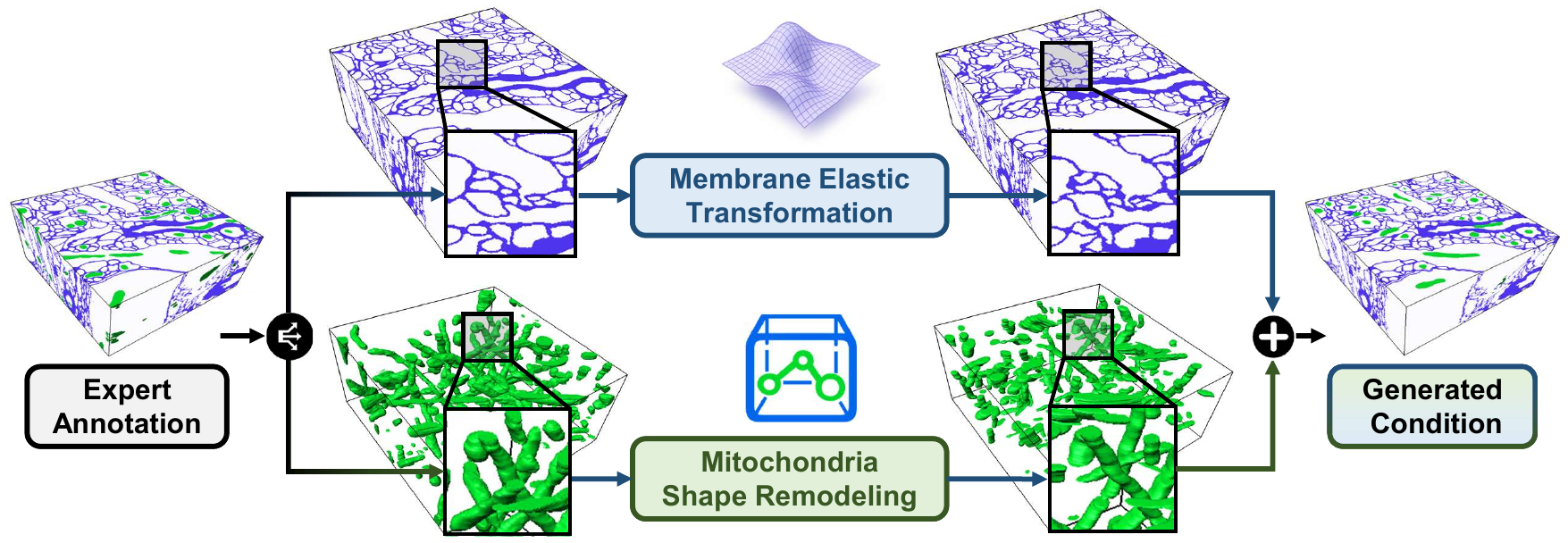}
	\vspace{-0.80cm}
	\caption{Illustration of the Biology-Guided Mask Remodeling.}
	\label{fig2_1}
	\vspace{-0.3cm}
\end{figure}
We observed that mitochondrial membranes can influence the identification of 
neuronal boundaries. Therefore, both neuronal boundaries and mitochondrial 
membranes are incorporated as conditional inputs, enabling the diffusion model 
to more effectively distinguish membrane types and generate EM images with 
higher fidelity, as shown in Fig.~\ref{fig2_1}.

We first construct a mitochondrial signature library by extracting 
boundary-based morphological descriptors from existing mitochondrial masks. For 
each mitochondrion, boundary contour points are extracted, and an equivalent 
ellipse is fitted using second-order central moments to compute its major axis 
length and orientation. Subsequently, we propose a selective elastic 
deformation strategy for neuronal membrane enhancement. Specifically, elastic 
deformation is applied to the binary membrane mask (0–1) and utilized as a 
conditional input for EM image synthesis. Unlike previous approaches that 
deform the entire image, our method perturbs only membrane structures. 
Experiments demonstrate that selective membrane deformation yields notable 
improvements in neuron segmentation, likely because local perturbations 
introduce realistic boundary variations while preserving global structural 
integrity.

Next, mitochondria are placed on the binary mask of the elastically deformed 
neuronal membrane. Since mitochondria typically appear in larger neuronal 
cross-sections, the top 10\% of neurons by volume are selected as candidate 
regions. Candidate neurons are randomly sampled, and ellipses are fitted to 
estimate their major axis lengths and orientations. Mitochondria from the 
signature library are then selected based on a predefined axis-length ratio 
relative to the corresponding neurons. During placement, 
each mitochondrion is aligned with the corresponding neuron’s orientation, 
followed by morphological refinement and boundary validation to ensure a 
biologically plausible distribution. This process generates a new mask from the 
original, which can then be used to synthesize new EM images.

\begin{table}[t]
	\centering
	\small
	\caption{Comparison of segmentation performance using different synthesis 
		methods for data augmentation across varying annotation ratios on the 
		AC3/AC4 datasets. ``Real images'' refer to training with manually 
		annotated 
		data only.}
	%\fontsize{9}{11}\selectfont
	\renewcommand{\arraystretch}{0.5}
	\setlength{\tabcolsep}{1.8pt}
	\begin{tabular}{c|c|cc|cc}
		\toprule
		\multicolumn{1}{c|}{\multirow{2}{*}{Annotate(\%)}} &  
		\multicolumn{1}{c|}{\multirow{2}{*}{Methods}} &
		\multicolumn{2}{c|}{Waterz} & 
		\multicolumn{2}{c}{Multicut} \\ 
		\cmidrule{3-6} 
		%		\multicolumn{2}{c|}{}   
		&                                            
		& $\mathrm{VI}\downarrow$    & $\mathrm{ARAND}\downarrow$          
		& $\mathrm{VI}\downarrow$    & $\mathrm{ARAND}\downarrow$ \\ 
		\midrule
		\multicolumn{1}{c}{\multirow{4}{*}{4\%}}
		%		\multirow{4}{*}{4\%}
		
		& \multicolumn{1}{|c|}{Real images}   
		& 1.623   
		& 0.209         
		& 1.437          
		& 0.189  \\ 
		
		\cmidrule{2-6} 
		
		& \multicolumn{1}{c|}{Pix2Pix}   
		& 1.779   
		& 0.222        
		& 1.461          
		& 0.141  \\ 
		
		& \multicolumn{1}{c|}{Med-DDPM}  
		& 1.493   
		& 0.170        
		& 1.421          
		& 0.206  \\
		
		& \multicolumn{1}{c|}{Ours}   
		& \textbf{1.376}   
		& \textbf{0.142}         
		& \textbf{1.277}          
		& \textbf{0.131} \\
		
		\midrule
		\multicolumn{1}{c}{\multirow{4}{*}{20\%}}
		%		\multirow{4}{*}{4\%}
		
		& \multicolumn{1}{|c|}{Real images}  
		& 1.150  
		& 0.117         
		& 1.186         
		& 0.133  \\ 
		
		\cmidrule{2-6} 
		
		& \multicolumn{1}{c|}{Pix2Pix}  
		& 1.185  
		& 0.135         
		& 1.342          
		& 0.218  \\ 
		
		& \multicolumn{1}{c|}{Med-DDPM}  
		& 1.147   
		& 0.122         
		& 1.195          
		& 0.133  \\
		
		& \multicolumn{1}{c|}{Ours}  
		& \textbf{1.115}   
		& \textbf{0.115}         
		& \textbf{1.114}          
		& \textbf{0.118}  \\
		
		\midrule
		\multicolumn{1}{c}{\multirow{4}{*}{100\%}}
		%		\multirow{4}{*}{4\%}
		
		& \multicolumn{1}{|c|}{Real images}   
		& 1.105   
		& 0.118         
		& \textbf{1.116}        
		& 0.111  \\ 
		
		\cmidrule{2-6} 
		
		& \multicolumn{1}{c|}{Pix2Pix}   
		& 1.451   
		& 0.161         
		& 1.769          
		& 0.290  \\ 
		
		& \multicolumn{1}{c|}{Med-DDPM}  
		& 1.106   
		& 0.106         
		& 1.148          
		& 0.107  \\
		
		& \multicolumn{1}{c|}{Ours}  
		& \textbf{1.102}   
		& \textbf{0.105}         
		& 1.135          
		& \textbf{0.105}  \\

		\bottomrule
	\end{tabular}
	\vspace{-0.4cm}
	\label{tab1}
\end{table}

\begin{table}[t]
	\centering
	\small
	\caption{Quantitative results of generated image quality.}
	\setlength{\tabcolsep}{42.0pt}
	\renewcommand{\arraystretch}{0.5}
	\begin{tabular}{c|c}
		\toprule
		\multicolumn{1}{c|}{Methods} &  
		\multicolumn{1}{c}{3D-FID$\downarrow$} \\ 
		
		\midrule      
		\multicolumn{1}{c|}{Pix2Pix}
		& 9.314   \\ 
		\multicolumn{1}{c|}{Med-DDPM}
		& 7.010   \\
		\multicolumn{1}{c|}{Ours}
		& \textbf{6.203}   \\

		\bottomrule
	\end{tabular}
	\vspace{-0.2cm}
	\label{tab-fid}
\end{table}

\section{Experiments}

\subsection{Datasets and Metrics}
We used the AC3 and AC4 datasets from the mouse somatosensory cortex 
\cite{AC2015}, imaged using scanning electron microscopy (SEM) at a resolution 
of $6 \times 6 \times 29~\mathrm{nm}^3$, with volumes of $256 \times 1024 
\times 1024$ and $100 \times 1024 \times 1024$, respectively. In our 
experiments, we used the AC4 volume for training and the first 100 slices of 
AC3 for testing. To assess the effectiveness of the proposed data augmentation 
strategy, we trained and evaluated neuron segmentation performance using 4\%, 
20\%, and 100\% of the available labeled data. We evaluated our method from two 
perspectives: assessing the quality of generated images and quantifying 
improvements in segmentation performance. To evaluate image quality, we adopted 
the widely used Fréchet Inception Distance (FID), extended to a 3D variant 
referred to as 3D-FID. Specifically, we pretrained a segmentation network on 
the AC3/AC4 datasets to extract feature representations and compute the 
distributional distance between real and generated images in the feature space, 
yielding the 3D-FID score. We evaluated neuron segmentation performance in EM 
images using two widely adopted metrics: Variation of Information (VI) and 
Adapted Rand Error (ARAND). Lower values for both metrics indicate better 
segmentation quality.

\subsection{Implementation Details}
Our experiments were conducted in two stages. In the first stage, the 
generative model was trained to synthesize new training samples, and in the 
second stage, the generated data were used to augment the training of the 
neuron segmentation network. Pix2Pix \cite{pixel} and Med-DDPM \cite{Medddpm} 
were used as baseline models, with their architectures adapted to 3D. Our model 
and Med-DDPM were trained for 10,000 iterations with a learning rate of $1 
\times 10^{-5}$, a batch size of 1, and input patch size of $8 \times 512 
\times 512$. All subsequent neuron segmentation experiments employed the 
Superhuman \cite{superhuman} model, which was trained for 200,000 iterations 
with a learning rate of $1 \times 10^{-4}$ and a batch size of 2. All models 
were trained on a single NVIDIA GeForce RTX 4090 GPU.

\begin{table}[t]
	\centering
	\small
	\caption{Comparison of segmentation performance with and without data 
		augmentation using different methods on the AC3/AC4 datasets.
	}
	\setlength{\tabcolsep}{0.5pt}
	\renewcommand{\arraystretch}{0.5}
	\begin{tabular}{cc|cc|cc}
		\toprule
		\multicolumn{2}{c|}{\multirow{2}{*}{Methods}} &  
		\multicolumn{2}{c|}{Waterz} &
		\multicolumn{2}{c}{Multicut} \\ 
		\cmidrule{3-6} 
		\multicolumn{2}{c|}{} &
		\multicolumn{2}{c|}{$\mathrm{VI}\downarrow$ \qquad 
			$\mathrm{ARAND}\downarrow$} &
		\multicolumn{2}{c}{$\mathrm{VI}\downarrow$ \qquad 
			$\mathrm{ARAND}\downarrow$} \\ 
		\midrule     
		\multicolumn{2}{l|}{Superhuman\cite{superhuman}}
		& 1.150 & 0.117 & 1.186 & 0.133 \\ 
		\multicolumn{2}{l|}{SwinUNETR\cite{SwinUNETR}}
		& 2.653 & 0.454 & 2.186 & 0.487 \\
		\multicolumn{2}{l|}{SegMamba\cite{SegMamba}}
		& 1.467 & 0.206 & 1.410 & 0.232 \\
		\midrule     
		\multicolumn{2}{l|}{Superhuman + Ours}
		& 1.115 & 0.115 & 1.114 & 0.118 
		\\ 
		\multicolumn{2}{l|}{SwinUNETR + Ours}
		& 2.294 & 0.375 & 1.921 & 0.419 \\
		\multicolumn{2}{l|}{SegMamba + Ours}
		& 1.387 & 0.187 & 1.269 & 0.177 \\

		\bottomrule
	\end{tabular}
	\vspace{-0.2cm}
	\label{tabUniversality}
\end{table}

\subsection{Comparison with State-of-the-art Methods}
We compared the quality of images generated by different methods and evaluated 
the corresponding changes in segmentation performance after data augmentation. 
All segmentation models were trained from scratch and employed the same 
post-processing pipelines: Waterz \cite{mala} and Multicut \cite{multicut}. For 
data augmentation using generated images, the ratio of augmented to original 
data was set to $1{:}1$.

\begin{figure}[t]
	\centering
	\includegraphics[width=0.48\textwidth]{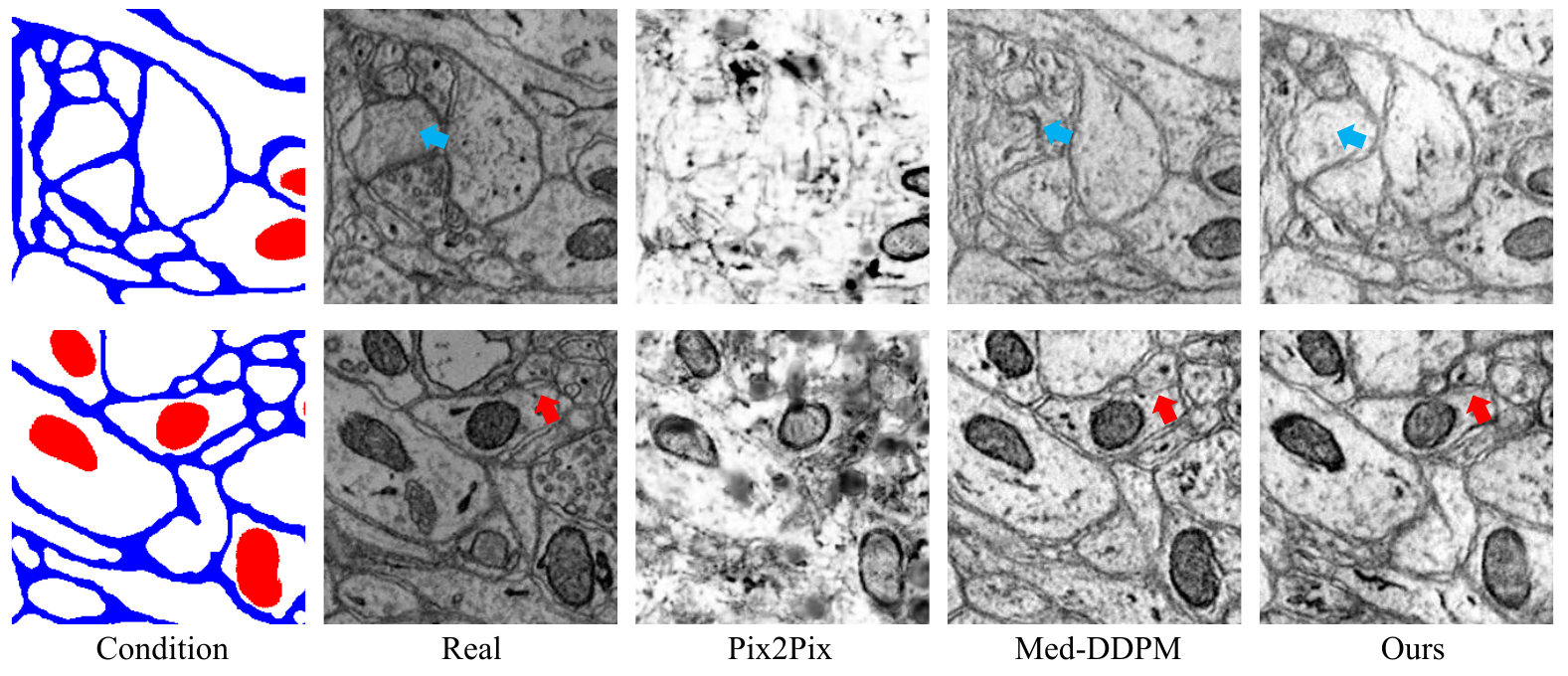} % Reduce the figure size so 
	%that it is slightly narrower than the column.
	\vspace{-0.8cm}
	\caption{Visualization of generated image quality. Blue arrows highlight 
		potential artifacts, while red arrows indicate possible membrane loss.}
	\label{fig3}
	\vspace{-0.2cm}
\end{figure}

\noindent\textbf{Generated image quality.}
We trained the generative model on 80 EM images of size $512 \times 512$ from 
the AC4 dataset and generated samples for the remaining 20 slices. 
Table~\ref{tab-fid} reports the 3D-FID scores of the three methods, where our 
approach achieves the best performance. Fig.~\ref{fig3} shows EM images 
generated by different methods under conditional guidance. Benefiting from the 
MSC module, our method provides more precise structural control.

\noindent\textbf{Segmentation Results.}
We conducted experiments on the AC3/AC4 datasets with three annotation 
ratios of 4\%, 20\%, and 100\%. Table~\ref{tab1} compares the segmentation 
performance obtained using only annotated data and that obtained using both 
generated and annotated data. Our data augmentation method consistently 
achieves superior performance across all annotation ratios. Specifically, with 
a 4\% annotation ratio, our augmentation method improves the ARAND metric by 
\textbf{32.1\%} and \textbf{30.7\%} under two distinct post-processing 
settings. To further assess the generality of our method, we conducted 
experiments on three representative segmentation frameworks: the CNN-based 
Superhuman \cite{superhuman}, the Transformer-based SwinUNETR \cite{SwinUNETR}, 
and the Mamba-based SegMamba \cite{SegMamba}. As shown in 
Table~\ref{tabUniversality}, 
with a 20\% annotation ratio, our data augmentation strategy consistently 
enhances the segmentation performance across all three models.

\begin{table}[t]
	\centering
	\small
	\caption{Ablation of MSC and RGM.}
	\setlength{\tabcolsep}{1.8pt}
	\renewcommand{\arraystretch}{0.5}
	\begin{tabular}{cc|cc|cc|c}
		\toprule
		\multicolumn{1}{c}{\multirow{2}{*}{MSC}} &  
		\multicolumn{1}{c|}{\multirow{2}{*}{RGM}} &
		\multicolumn{2}{c|}{Waterz} &
		\multicolumn{2}{c|}{Multicut} &
		\multicolumn{1}{c}{\multirow{2}{*}{3D-FID$\downarrow$}}\\ 
		\cmidrule{3-6} 
		\multicolumn{2}{c|}{} &
		\multicolumn{2}{c|}{$\mathrm{VI}\downarrow$ \quad 
			$\mathrm{ARAND}\downarrow$} &
		\multicolumn{2}{c|}{$\mathrm{VI}\downarrow$ \quad 
			$\mathrm{ARAND}\downarrow$} \\ 
		\midrule     
		\multicolumn{1}{c}{}  & \multicolumn{1}{c|}{}  
		& 1.765 & 0.180 & 1.577 & 0.174 & 6.838\\ 
		\multicolumn{1}{c}{\checkmark}  & \multicolumn{1}{c|}{} 
		& 1.649 & 0.164 & 1.565 & 0.172 & 6.734\\  
		\multicolumn{1}{c}{}  & \multicolumn{1}{c|}{\checkmark}  
		& 1.587 & 0.162 & 1.425 & 0.140 & 6.603\\ 
		\multicolumn{1}{c}{\checkmark}  & 
		\multicolumn{1}{c|}{\checkmark}  
		& \textbf{1.376} & \textbf{0.142} & \textbf{1.277} & 
		\textbf{0.131} & \textbf{6.203} \\ 
		\bottomrule
	\end{tabular}
	\label{tab4}
	\vspace{-0.3cm}
\end{table}

\begin{table}[t]
	\centering
	\small
	\caption{Ablation Study on Mask Remodeling.}
	\setlength{\tabcolsep}{2.5pt}
	\renewcommand{\arraystretch}{0.5}
	\begin{tabular}{cc|cc|cc}
		\toprule
		\multicolumn{2}{c|}{\multirow{2}{*}{Methods}} &  
		\multicolumn{2}{c|}{Waterz} &
		\multicolumn{2}{c}{Multicut} \\ 
		\cmidrule{3-6} 
		\multicolumn{2}{c|}{} &
		\multicolumn{2}{c|}{$\mathrm{VI}\downarrow$ \qquad 
			$\mathrm{ARAND}\downarrow$} &
		\multicolumn{2}{c}{$\mathrm{VI}\downarrow$ \qquad 
			$\mathrm{ARAND}\downarrow$} \\ 
		\midrule     
		\multicolumn{2}{l|}{No Remodeling}
		& 1.427 & 0.156 & 1.331 & 0.138 \\ 
		\multicolumn{2}{l|}{Neuron-only}
		& 1.427 & 0.143 & \textbf{1.252} & 0.137 \\
		\multicolumn{2}{l|}{Mito-only}
		& 1.415 & 0.151 & 1.313 & \textbf{0.131} \\
		\multicolumn{2}{l|}{Ours}
		& \textbf{1.376} & \textbf{0.142} & 1.277 & \textbf{0.131} \\ 
		\bottomrule
	\end{tabular}
	\label{tab5}
	\vspace{-0.2cm}
\end{table}

\subsection{Ablation Studies and Analysis}
We conducted ablation experiments using 4\% of the labeled data to evaluate the 
effectiveness of the MSC and RGM modules. Table~\ref{tab4} summarizes the 
experimental results under different module configurations. In terms of both 
generated image quality and segmentation performance, the combined use of MSC 
and RGM achieves the best results. This finding suggests that generating more 
accurate neuronal imagery requires both multi-scale conditional guidance and 
global spatial connectivity modeling among voxels.

We further validate the effectiveness of Biology-Guided Mask Remodeling. 
Table~\ref{tab5} presents the experimental results for four settings: 
no remodeling, remodeling only the neuronal membrane, remodeling only the 
mitochondrial mask, and remodeling both simultaneously. 
The results indicate that jointly remodeling the neuronal membrane and 
mitochondrial mask produces more diverse training data, thereby achieving the 
best segmentation performance.

\section{Conclusion}
This paper proposes a diffusion-based data augmentation framework for neuron 
segmentation, which synthesizes new image–label pairs to enhance model 
training. Experiments demonstrate that our method effectively improves neuron 
segmentation performance.

\section{Compliance with ethical standards}
This research study was conducted retrospectively using mouse subject data made 
available in open access \cite{AC2015}. Ethical approval was not required 
as confirmed by the license attached with the open access data.

\section{Acknowledgements}
This work was supported by the Beijing Natural Science Foundation 
(No.~5254042), the Brain Science and Brain-like Intelligence Technology 
-- National Science and Technology Major Project (2022ZD0211900, 
2022ZD0211902), and the National Natural Science Foundation of China 
(No.~32171461).

\bibliographystyle{IEEEbib}
\bibliography{reference}

\end{document}